\documentclass{bioinfo}
\copyrightyear{2013}
\pubyear{2013}
\usepackage{subfigure}
\usepackage{listings}
\begin{document}

\firstpage{1}

\title[RRF]{Guided Random Forest in the RRF Package}
\author[Houtao Deng]{Houtao Deng\,}
\address{Intuit, Mountain View, CA 94043, USA.\\}

\history{}
\editor{}

\maketitle

\pagestyle{plain}

\begin{abstract}

\section{Summary:}
  Random Forest (RF) is a powerful supervised learner and has been popularly used in many applications such as bioinformatics.
  In this work we propose the guided random forest (GRF) for feature selection.

  Similar to a feature selection method called guided regularized random forest (GRRF), GRF is built using the importance scores from an ordinary RF. However, the trees in GRRF are built sequentially, are highly correlated and do not allow for parallel computing, while the trees in GRF are built independently and can be implemented in parallel.

  Experiments on 10 high-dimensional gene data sets show that, with a fixed parameter value (without tuning the parameter), RF applied to features selected by GRF outperforms RF applied to all features on 9 data sets and 7 of them have significant differences at the 0.05 level. Therefore, both accuracy and interpretability are significantly improved. GRF selects more features than GRRF, however, leads to better classification accuracy.

  Note in this work the guided random forest is guided by the importance scores from an ordinary random forest, however, it can also be guided by other methods such as human insights (by specifying $\lambda_i$).

  GRF can be used in ``RRF" v1.4 (and later versions), a package that also includes the regularized random forest methods.

\section{Availability:}
The RRF package is freely distributed under GNU General Public License (GPL) and is available from CRAN (http://cran.r-project.org/), the official R package archive.

\section{Contact:} \href{hdeng3@asu.edu}{hdeng3@asu.edu}
\end{abstract}


\section{Introduction}
Random forest (RF) \citep{breiman2001,Andy2002} has been widely used in many fields including bioinformatics applications \citep{riddick2011predicting,yuan2012}. RF is able to handle mixed categorical and numerical features, multiple classes, are insensitive to the scale of features, and have been considered as a powerful supervised learner.

RF can provide importance scores of features to understand the contribution of each feature. However, there can be a huge number of features for high-dimensional problems (all the gene data sets considered in our experiments have more than 1000 features), and it is
challenging to investigate the importance scores from thousands of features. Therefore, it is desirable to develop a feature selection algorithm for RF. .

The guided regularized random forest (GRRF) proposed by \cite{deng2012gene} uses the importance scores from an RF built on the complete training data to complement the information gain in a local node. However, the trees in GRRF can be highly correlated and GRRF can not be built in parallel \citep{deng2012gene}.

The guided random forest (GRF), proposed in this work, is a solution of the issues mentioned above. GRF is guided by the importance scores
from an RF, and each tree in GRF is built independently from another tree. Experiments on 10 gene data sets show conclusive results that
GRF uses many fewer features than RF, and RF applied to features selected by GRF is more accurate than RF.

\begin{methods}
\section{Methods}

Let $gain(X_i)$ denote the Gini information gain of using a feature $X_i$ to split a tree node.
The key idea of GRF is weighting $gain(X_i)$ using the importance scores from an RF.

 \begin{equation}
gain_G(X_i) =  \lambda_i\ gain(X_i)
\end{equation}
where $\lambda_i$ is calculated as
\begin{equation}\label{eq:lambda}
 \lambda_i = 1-\gamma + \gamma\ \frac{Imp_i}{Imp^*}\\
\end{equation}
where $Imp_i$ is the importance score of $X_i$ from an RF,
$Imp^*$ is the maximum importance score, $\frac{Imp_i}{Imp^*}\in [0,1]$ is the normalized importance score,
and $\gamma\in[0,1]$ controls the weight of the importance scores from RF.
It can be seen that, 
 features with smaller importance scores are penalized more in GRF, and the penalty increases as $\gamma$ increases (GRF becomes RF when $\gamma$ = 0). In this work I use the maximum penalty (i.e., $\gamma=1$), in order to use a small number of features in GRF. So $gain_G(X_i)$ becomes
\begin{equation}
\frac{Imp_i}{Imp^*}\ gain(X_i)
\end{equation}
Note the key difference between GRF and GRRF is that the features used in previous trees have an impact on the current tree for GRRF, but does not have any impact for GRF. \emph{The features used in a GRRF model are expected to be {relevant and non-redundant}, while the features used in a GRF model are expected to be relevant, but not necessarily non-redundant.}
\end{methods}

\section{Examples and Results}
Code \ref{rcode} shows an example of using GRF ($\gamma=1$) for feature selection. In the code, a classification data set with 500 features
is simulated, and only 2 features are relevant to the class. While RF uses all the features and misclassifies 54 out of 250 instances, RF uses 196 features selected by GRF and misclassifies 34.

\definecolor{mygreen}{rgb}{0,0.6,0}
\lstset{
  belowcaptionskip=1\baselineskip,
  frame=single,
  language=R,
  basicstyle=\scriptsize\ttfamily,
  commentstyle=\color{blue},
}
\renewcommand{\lstlistingname}{Code}

\begin{table}[t]
\tiny
\centering
\caption{Error rates of GRF-RF (RF applied to the feature subset
selected by GRF), GRF (as a classifier), RF, GRRF (as a classifier), GRRF-RF (RF applied to the feature subset
selected by GRRF), averaged over 100 runs. Let ``$\circ$" or ``$\bullet$" denote a significant difference between a method and GRF-RF at the 0.05 level, according to the paired t-test.
Particularly, ``$\circ$" or ``$\bullet$" standards for a higher or lower error rate of a method compared to GRF. \label{Tab:01}}
\vspace{-0.2cm}
{
\begin{tabular}{cccccccccc}
\hline
           & {\bf GRF-RF} &  {\bf GRF} &     {\bf } &   {\bf RF} &     {\bf } & {\bf GRRF} &     {\bf } & {\bf GRRF-RF} &     {\bf } \\
\hline
adenocarcinoma &      0.156 &      0.159 &            &      0.155 &            &      0.187 &   $\circ$ &      0.168 &   $\circ$ \\

     brain &      0.144 &      0.168 &   $\circ$ &      0.178 &   $\circ$ &      0.263 &   $\circ$ &      0.216 &   $\circ$ \\

breast.2.class &      0.348 &      0.348 &            &      0.368 &   $\circ$ &      0.355 &            &      0.346 &            \\

breast.3.class &      0.397 &      0.399 &            &      0.417 &   $\circ$ &      0.395 &            &      0.390 &            \\

     colon &      0.149 &      0.142 &   $\bullet$ &      0.162 &   $\circ$ &      0.198 &   $\circ$ &      0.171 &   $\circ$ \\

  leukemia &      0.044 &      0.037 &            &      0.061 &   $\circ$ &      0.106 &   $\circ$ &      0.088 &   $\circ$ \\

  lymphoma &      0.007 &      0.008 &            &      0.007 &            &      0.091 &   $\circ$ &      0.068 &   $\circ$ \\

       nci &      0.313 &      0.331 &   $\circ$ &      0.319 &            &      0.463 &   $\circ$ &      0.382 &   $\circ$ \\

  prostate &      0.086 &      0.093 &   $\circ$ &      0.108 &   $\circ$ &      0.101 &   $\circ$ &      0.087 &            \\

     srbct &      0.024 &      0.029 &   $\circ$ &      0.031 &   $\circ$ &      0.096 &   $\circ$ &      0.050 &   $\circ$ \\
\hline
win-lose-tie &    {\bf -} & {\bf 2-7-1} &     {\bf } & {\bf 1-9-0} &     {\bf } & {\bf 1-9-0} &     {\bf } & {\bf 2-8-0} &     {\bf } \\
\hline
\end{tabular}

}{}
\end{table}

\vspace{5.0cm}
\begin{lstlisting}[caption={Feature Selection and Classification with GRF},label={rcode}]

library(RRF) # load the RRF package
set.seed(1) # fix the random seed.
# simulate classification data set
X <- matrix(runif(500*500, min=-1, max=1), ncol=500)
# class is only relevant to feature 1 and 21
Y <- (X[,1]) + (X[,21])
ix <- which(Y>quantile(Y, 1/2));
Y <- Y*0-1; Y[ix] <- 1 #assign class -1 and 1
#split data into training and testing sets
trainIx <- 1:250
trainX <- X[trainIx,]; trainY <- Y[trainIx]
testX <- X[-trainIx,]; testY <- Y[-trainIx]
# build an ordinary RF on the training instances
# note 'flagReg=1' results in a regularized RF
RF <- RRF(trainX,flagReg=0, as.factor(trainY))
print(length(RF$feaSet)) #500 features used
imp <- RF$importance[,"MeanDecreaseGini"]
impRF <- imp/max(imp) # normalization
# build a GRF with gamma = 1
# note the difference between GRF and RF is that
# 'coefReg' is related to impRF in GRF, while
# it is constant for all variables in RF.
gamma <- 1
coefReg <- (1-gamma) + gamma*impRF
GRF <- RRF(trainX,as.factor(trainY),
           flagReg=0,coefReg=coefReg)
print(length(GRF$feaSet)) #196 features used
GRF_RF <- RRF(trainX[,GRF$feaSet],
             flagReg=0, as.factor(trainY))
# test RF and GRF on the testing instances
pred <- predict(RF,testX)#predict using RF
# 54 instances misclassified
print(length(which(pred != testY)))
# predict using GRF's features
pred <- predict(GRF_RF,testX[,GRF$feaSet])
# 34 instances misclassified
print(length(which(pred != testY)))


\end{lstlisting}

\vspace{0cm}

In addition, I applied GRF-RF (RF applied to the feature subset selected by GRF), GRF, GRRF ($\gamma=0.1$) and GRRF-RF (RF applied to the feature subset selected by GRRF), all with 1000 trees, to 10 gene data sets used in \cite{diaz2006gene,deng2012gene}. The references of the data sets are provided in a supplementary file to save space. I obtained the average error rates and average number of features for each method using the same procedure as \cite{deng2012gene}, i.e., calculated from 100 replicates of training/testing splits with a ratio of 2:1. The results of RF and GRRF-RF are slightly different from the results of \cite{deng2012gene} due to randomness. Table~\ref{Tab:01} shows the average error rates of different methods. GRF-RF outperforms RF on 9 data sets, 7 of them have significant differences at the 0.05 level. The advantage of GRF-RF over GRRF and GRRF-RF is also clear. GRF-RF also outperform GRF, and therefore applying RF to features selected by GRF is better than GRF as a classifier.

\vspace{-0.5cm}
\begin{table}[!h]
\scriptsize
\centering
\vspace{0.4cm}\caption{The number of instances, classes and features of the data sets, and the number of features used in GRF and RF. \label{Tab:Fea}}
\vspace{-0.2cm}
\begin{tabular}{cccccc}
\hline
    {\bf } & {\bf Instances} & {\bf Classes} & {\bf Features} &  {\bf GRF} &   {\bf RF} \\
\hline
adenocarcinoma &         76 &          2 &       9868 &        472 &       2143 \\

     brain &         42 &          5 &       5597 &        397 &       2621 \\

breast.2.class &         77 &          2 &       4869 &        385 &       2654 \\

breast.3.class &         95 &          3 &       4869 &        421 &       3518 \\

     colon &         62 &          2 &       2000 &        234 &       1291 \\

  leukemia &         38 &          2 &       3051 &        228 &        788 \\

  lymphoma &         62 &          3 &       4026 &        295 &       1226 \\

       nci &         61 &          8 &       5244 &        444 &       3877 \\

  prostate &        102 &          2 &       6033 &        414 &       2716 \\

     srbct &         63 &          4 &       2308 &        262 &       1630 \\
\hline
\end{tabular}
\end{table}

\vspace{-0.2cm}
Table~\ref{Tab:Fea} summarizes the data sets and shows the number of features used in different models. RF uses a subset of features in the model, and GRF uses a even smaller number of features in the model. 

\section{Conclusions}
The guided random forest (GRF) is proposed here for feature selection, particularly, for gene classification in this work.  Experiments show that GRF-RF not only significantly outperforms RF in accuracy performance, but also uses many fewer features in the model. In this work I discuss the advantages of GRF for high-dimensional gene data sets. It may also be valuable to find other cases where GRF has advantages over other methods, with the option of tuning the parameter $\gamma$ in Equation (\ref{eq:lambda}) (fixed as 1 here). Furthermore, in this work, $\lambda_i$ is determined by the importance score of feature $X_i$ from an ordinary random forest. However, $\lambda_i$ can be specified by other ways too, e.g., F-score or human knowledge.

\bibliographystyle{natbib}
\bibliography{RRF}

\begin{thebibliography}{}

\bibitem[Breiman(2001)Breiman]{breiman2001}
Breiman, L. (2001).
\newblock Random forests.
\newblock {\em Machine Learning\/}, {\bf 45}(1), 5--32.

\bibitem[Deng and Runger(2013)Deng and Runger]{deng2012gene}
Deng, H. and Runger, G. (2013).
\newblock Gene selection with guided regularized random forest.
\newblock {\em Pattern Recognition\/}.
\newblock to appear.

\bibitem[D{\'\i}az-Uriarte and De~Andres(2006)D{\'\i}az-Uriarte and
  De~Andres]{diaz2006gene}
D{\'\i}az-Uriarte, R. and De~Andres, S. (2006).
\newblock Gene selection and classification of microarray data using random
  forest.
\newblock {\em BMC bioinformatics\/}, {\bf 7}(1), 3.

\bibitem[Liaw and Wiener(2002)Liaw and Wiener]{Andy2002}
Liaw, A. and Wiener, M. (2002).
\newblock Classification and regression by randomforest.
\newblock {\em R News\/}, {\bf 2}(3), 18--22.

\bibitem[Riddick {\em et~al.}(2011)Riddick, Song, Ahn, Walling, Borges-Rivera,
  Zhang, and Fine]{riddick2011predicting}
Riddick, G., Song, H., Ahn, S., Walling, J., Borges-Rivera, D., Zhang, W., and
  Fine, H.~A. (2011).
\newblock Predicting in vitro drug sensitivity using random forests.
\newblock {\em Bioinformatics\/}, {\bf 27}(2), 220--224.

\bibitem[Yuan {\em et~al.}(2012)Yuan, Xu, Xu, Ball, and Liang]{yuan2012}
Yuan, Y., Xu, Y., Xu, J., Ball, R.~L., and Liang, H. (2012).
\newblock Predicting the lethal phenotype of the knockout mouse by integrating
  comprehensive genomic data.
\newblock {\em Bioinformatics\/}, {\bf 28}(9), 1246--1252.

\end{thebibliography}

\end{document}